\documentclass[journal]{IEEEtran}

\usepackage{cite}
\usepackage[protrusion=true,expansion=true]{microtype}
\usepackage{amsmath,amssymb,amsfonts}
\usepackage{graphicx}
\usepackage{booktabs}
\usepackage{array,tabularx}
\newcolumntype{L}{>{\raggedright\arraybackslash}X}
\newcommand{\TableSetup}{%
  \centering\footnotesize
  \setlength{\tabcolsep}{3pt}%
  \renewcommand{\arraystretch}{1.0}%
  \setlength{\heavyrulewidth}{0.6pt}%
  \setlength{\lightrulewidth}{0.4pt}%
  \setlength{\cmidrulewidth}{0.3pt}%
  \setlength{\aboverulesep}{1pt}%
  \setlength{\belowrulesep}{1pt}}
\newcommand{\TableHead}[1]{\shortstack[c]{\bfseries #1}}
\newcommand{\TableNotes}[1]{%
  \par\vspace{2pt}%
  \noindent\parbox{\linewidth}{\footnotesize\raggedright\textit{Notes.}~#1\par}}

\usepackage{textcomp}
\usepackage{xcolor}
\usepackage[hidelinks,breaklinks=true]{hyperref}

\graphicspath{{figs/}}

\makeatletter
\g@addto@macro\normalsize{%
  \setlength{\abovedisplayskip}{5pt plus 2pt minus 3pt}%
  \setlength{\belowdisplayskip}{5pt plus 2pt minus 3pt}%
  \setlength{\abovedisplayshortskip}{2pt plus 2pt}%
  \setlength{\belowdisplayshortskip}{3pt plus 2pt minus 2pt}}
\makeatother
\newcommand{\ie}{i.e.,\ }

\newcommand{\Real}{\mathbb{R}}
\newcommand{\Sph}{\mathbb{S}}
\newcommand{\Model}{\mathcal{M}}
\newcommand{\Obs}{\mathcal{Q}}
\newcommand{\Obsf}{\mathcal{Q}_{\mathrm{f}}}
\newcommand{\Sim}{\mathrm{Sim}(3)}
\newcommand{\R}{\mathbf{R}}
\newcommand{\tvec}{\mathbf{t}}
\newcommand{\desc}{\mathbf{d}}
\newcommand{\p}{\mathbf{p}}
\newcommand{\q}{\mathbf{q}}
\newcommand{\x}{\mathbf{x}}

\newcommand{\rmsr}{\rho}
\newcommand{\gscore}{\mathcal{G}}
\newcommand{\dis}{\mathcal{D}}
\DeclareMathOperator*{\argmin}{arg\,min}

\DeclareMathOperator{\softplus}{softplus}
\DeclareMathOperator{\median}{median}

\begin{document}

\title{Landmark-Free and Scale-Blind Multi-View Registration of Preoperative
Liver Models to Intraoperative Surfaces}

\author{Jiaming~Feng,
        Gerardo~Loza-Galindo,
        Shahid~Farid,
        and~Sharib~Ali%
\thanks{Corresponding author: Sharib Ali. (email: s.s.ali@leeds.ac.uk)}%
\thanks{Jiaming Feng, Gerardo Loza-Galindo, and Sharib Ali are with the
AI in Medicine and Surgery Group, School of Computer Science,
University of Leeds, Leeds LS2 9JT, United Kingdom
(e-mail: vqkc0507@leeds.ac.uk; G.E.LozaGalindo@leeds.ac.uk;
s.s.ali@leeds.ac.uk).}%
\thanks{Shahid Farid is with the Department of HPB and Transplant Surgery,
St.~James's University Hospital, Leeds, United Kingdom
(e-mail: s.farid@nhs.net).}%
\thanks{This work was supported by the Engineering and Physical Sciences Research Council [grant number UKRI914].}%
}

\bstctlcite{IEEEexample:BSTcontrol}
\maketitle

\begin{abstract}
Augmented-reality guidance in laparoscopic liver resection requires aligning preoperative 3D liver models from CT/MRI with intraoperative surfaces. Monocular reconstruction in most cases cannot guarantee scale, albeit rigid-motion registration requires scale as an input. Our proposed framework ``SimCompose'' estimates the full similarity transformation $\Sim$, including scale, without relying on landmarks and applies the same transformation across multiple views with known relative geometry.
However, reliable detection of registration failures remains challenging in surgical settings, where ground-truth alignment is generally unavailable.
For this, we propose a geometry-based selector that evaluates candidate transformations without requiring ground truth. Large disagreement among these transformations is used to flag potentially unreliable registrations.
Finally, we propose to use a non-rigid refinement based on a low-dimensional deformation model from Neo-Hookean simulations in the Simulation Open Framework Architecture (SOFA), without additional network training.
SimCompose is trained exclusively on synthetic partial views. Its encoder processes point clouds normalised to a unit root-mean-square radius, without access to metric scale, and filters each view before multi-view fusion.
Evaluation on unseen synthetic data (300 single-view pairs per run; 100 three-view registrations) shows that multi-view integration and candidate selection increase the success rate under stringent $\Sim$ accuracy criteria from \(38.4\%\) for a single-view, single-method configuration to \(77\%\). Existing learning-based baselines do not satisfy these criteria even when provided with the exact scale.
On five in-vivo patients with $92$ keyframes, the rigid registration stage reduces surface error more than three-fold relative to learning-based baselines with external scale information.

\end{abstract}


\begin{figure}[t]
\centering
\includegraphics[width=\columnwidth]{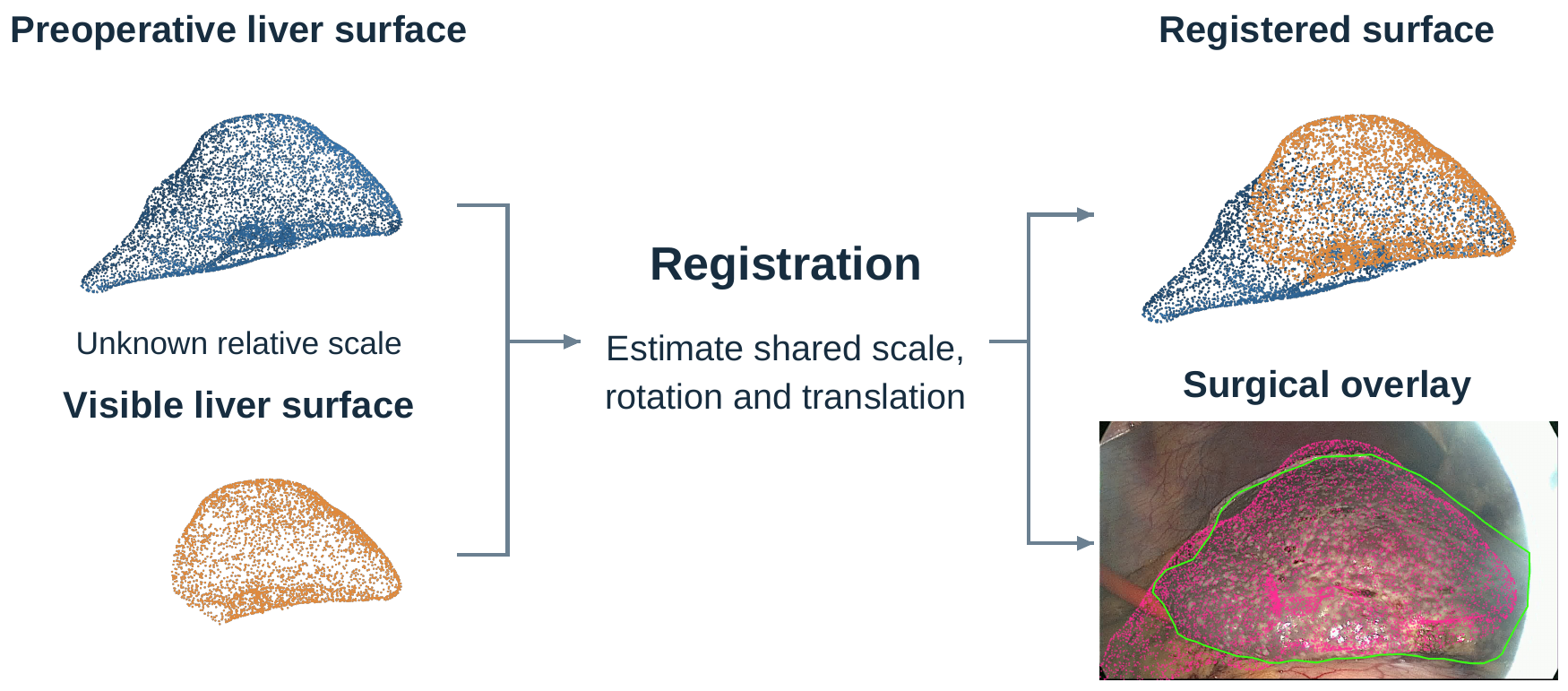}
\caption{The registration task. The complete preoperative liver surface is blue; partial laparoscopic observations with surrounding clutter are orange. We estimate one scale, rotation and translation to align the model to observations sharing a coordinate frame and relative scale, without landmarks or supplied metric scale. Non-rigid refinement is optional. Right: surgical overlay, with the transformed model in magenta and the annotated liver boundary in green.}
\label{fig:task}
\end{figure}

\section{Introduction}\label{sec:intro}

\IEEEPARstart{L}{aparoscopic} liver surgery provides a narrow and oblique view of the liver that is opaque and textureless. Augmented-reality (AR) guidance aligns a patient-specific preoperative extracted 3D model from the computed tomography (CT) or magnetic resonance imaging (MRI) scans with the laparoscopic view to locate subsurface tumours and vessels~\cite{nicolau2011augmented, modrzejewski2019invivo} (Fig.~\ref{fig:task}). Registration errors are critical for achieving AR guidance for accurate liver resection, usually measured in millimetres. While the literature suggests liver silhouettes, the falciform ligament, and the ridge as defined anatomical landmarks, these are highly inaccurate and variable because they require human expert labels and are usually thin, curve-like structures~\cite{ali2025p2ilf, koo2022automatic}. Landmark-based registration methods align anatomical curves with the model in 3D--2D~\cite{ali2025p2ilf, koo2022automatic, labrunie2023automatic, espinel2021combining, robu2018global}; however, these approaches cannot guarantee the optimal registration accuracy needed in tumour resections. Therefore, a framework that does not depend on anatomical landmarks is needed. Additionally, landmark-free surface registration methods such as Self-P2IR~\cite{zhou2025selfp2ir}, which align reconstructed surfaces in 3D--3D, still lack a reliable scale estimate.

In the literature, monocular Depth-Anything-3~\cite{lin2025depthanything3} predicts depth with scale and shift ambiguity; VGGT~\cite{wang2025vggt} recovers a shared non-metric scene scale. A uniform depth scale multiplies the reconstructed coordinates and can be absorbed into spatial scaling, whereas a depth offset moves points along their viewing rays and can change surface shape. Estimating one scale therefore addresses only the first ambiguity. GeoTransformer~\cite{qin2023geotransformer}, Lepard~\cite{li2022lepard} and PREDATOR~\cite{huang2021predator} estimate $\mathrm{SE}(3)$ transformations: rotation and translation without scale. Similarly, Self-P2IR~\cite{zhou2025selfp2ir} supplies scale from the principal-axis ratio of the visible intraoperative and complete preoperative surfaces, which changes with visibility. Beyond scale, registering frames independently discards shared multi-view constraints. Fusion requires common coordinates and consistent relative scale; extrinsics alone cannot reconcile independently scaled reconstructions.

Similarly, most of the landmark-free approaches are based on matching reconstructed surfaces, but limited coverage can leave several plausible transformations consistent with the same observation.

P2ILF, the preoperative-to-intraoperative laparoscopic fusion challenge~\cite{ali2025p2ilf}, made this approach reproducible, but sparse and often occluded landmarks can yield unrecognised pose errors. Self-P2IR~\cite{zhou2025selfp2ir} instead aligns reconstructed surfaces in 3D--3D.

A partial view of a smooth, nearly symmetric surface may fit several model regions, motivating comparison of multiple candidate transformations. SimCompose (Fig.~\ref{fig:method}) addresses these issues through the following contributions:
\begin{itemize}
    \item [1] a scale-blind registration method that jointly estimates scale, rotation and translation;
    \item [2] per-view point cloud filtering based on overlap and multi-view fusion of point cloud and its candidate selection without ground-truth labels;
    \item[3]  a candidate disagreement score for screening registration failures;
    \item [4] a rendered multi-view dataset with known camera geometry for testing silhouette consistency to tackle views excluded from registration;
    \item [5] evaluation on synthetic data and in-vivo surgical data to establish pose accuracy, surface agreement, and failure modes.
\end{itemize}

After a similarity transformation is selected, we refine the liver shape using a deformation model built from Neo-Hookean simulations in SOFA (Simulation Open Framework Architecture)~\cite{faure2012sofa, marchesseau2010fast}. We measure pose accuracy on synthetic observations from meshes not used to train the encoder, using the known registration transforms as ground truth. Rendered views complement this evaluation by testing whether one registration remains consistent across viewpoints. We also assess surface alignment on the in-vivo P2I-LReg dataset~\cite{zhou2025selfp2ir}, with Self-P2IR's rigid core as the main baseline. Because relative camera poses are unavailable, we register each in-vivo frame separately.

\section{Related Work}\label{sec:related}

Sparse anatomical features provide landmarks for aligning liver models~\cite{labrunie2023automatic,koo2022automatic,espinel2021combining,ozgur2018preoperative,robu2018global}, whereas patient-specific biomechanics constrain deformation~\cite{plantefeve2016patient,marchesseau2010fast,heiselman2024image,rucker2014mechanics,suwelack2014physics,haouchine2013image,pfeiffer2020nonrigid}. Fiducials and operator interaction supply alignment constraints~\cite{adagolodjo2018marker,modrzejewski2019invivo,nicolau2011augmented}. P2ILF~\cite{ali2025p2ilf} consolidates landmark-driven 3D--2D registration, while Self-P2IR~\cite{zhou2025selfp2ir} aligns monocular surfaces in 3D. Its complete system adds a deformation network trained with self-supervision within the cohort's cross-validation folds; we assess both its released rigid core (the rigid stage without this network) and its complete system.

Point-cloud backbones~\cite{wang2019dgcnn} and convolutional descriptors~\cite{zeng20173dmatch,choy2019fcgf,bai2020d3feat} extract features for matching. Overlap prediction~\cite{huang2021predator} identifies potentially shared regions, while superpoint attention~\cite{qin2023geotransformer}, position encoding and rotation-invariant representations~\cite{li2022lepard,yu2023roitr} provide geometric context. Sinkhorn matching~\cite{sarlin2020superglue} assigns soft correspondences, and non-rigid extensions~\cite{li2022ndp} represent deformation. Rigid estimators still assume supplied scale, which monocular reconstructions do not provide.

We combine learned correspondences with classical $\Sim$ estimation~\cite{rusu2009fpfh,fischler1981ransac,umeyama1991least,besl1992icp,chen1992object,myronenko2010cpd,yang2021teaser,chen2022sc2pcr} to recover scale, rotation and translation. Classical methods also provide supplementary matches and baselines with comparable computational effort.
\section{Rendered Data and Reconstruction Front Ends}\label{sec:blender}

Because the true alignment between the preoperative model and reconstructed scene is unknown, our rendered dataset uses virtual cameras with known poses to test cross-view consistency, \ie whether one registration agrees with the segmented liver in every view. The Visual Geometry Grounded Transformer (VGGT)~\cite{wang2025vggt}, a multi-view reconstruction network, recovers static laparoscopic scenes from P2ILF videos. The 3D graphics software Blender then renders a $200$-frame camera trajectory with exact intrinsics, extrinsics and depth. Backprojecting the rendered depth gives partial surface clouds; nearest-vertex correspondence propagates one liver segmentation of the reconstructed scene cloud across frames. We evaluate seven such rendered cases (hereafter, Blender cases), six of which come from patients in the in-vivo P2I-LReg cohort~\cite{zhou2025selfp2ir}. The preoperative meshes of two of these cases were in the training data of our encoders and of the previously trained weights used to initialise them (warm starts); the meshes of the other five were not used in encoder training.

Only the camera moves, so the shared reconstruction provides common coordinates and relative scale without inter-frame organ deformation. We estimate one transformation from three anchor views, the views used for fitting, and assess it in eight held-out views by silhouette Dice. Dice is twice the intersection area of the projected-model and segmented-liver silhouettes divided by their summed areas, reported on a $0$--$100$ scale. The held-out views can expose anchor fits that match the anchor silhouettes but have mirrored or flattened depth.

To select the reconstruction front ends, \ie the networks that turn laparoscopic video into point clouds, we benchmark predicted depth against structured-light depth from the Stereo Correspondence and Reconstruction of Endoscopic Data (SCARED) dataset~\cite{allan2021scared} on valid pixels with true depth $1<Z<150$\,mm. On $25$ SCARED keyframes with structured-light ground truth, we fit scale and offset in each predictor's depth or disparity (inverse-depth) space and measure the absolute relative error (AbsRel), the mean of per-pixel absolute errors normalised by their true depths. Among seven predictors, monocular Depth-Anything-3~\cite{lin2025depthanything3} has the lowest AbsRel, $0.0576$, versus $0.0591$ for the endoscopic monocular depth network AF-SfMLearner~\cite{shao2022af}. With true focal length and no alignment, metric-depth predictors applied zero-shot, \ie without training on these data, give AbsRel $5.14$--$24.47$. For multi-view reconstruction on $12$-frame sequences from $5$ keyframes, per-frame median-scale alignment gives AbsRel $0.043$ for VGGT~\cite{wang2025vggt}, $0.061$ for Endo3R~\cite{guo2025endo3r} and $0.123$ for DUSt3R~\cite{wang2024dust3r}. AF-SfMLearner and Endo3R were trained on the SCARED sequences used here. These alignments use ground-truth depth and serve only this benchmark.
\section{Method}\label{sec:method}

\begin{figure*}[t]
\centering
\includegraphics[width=0.96\textwidth]{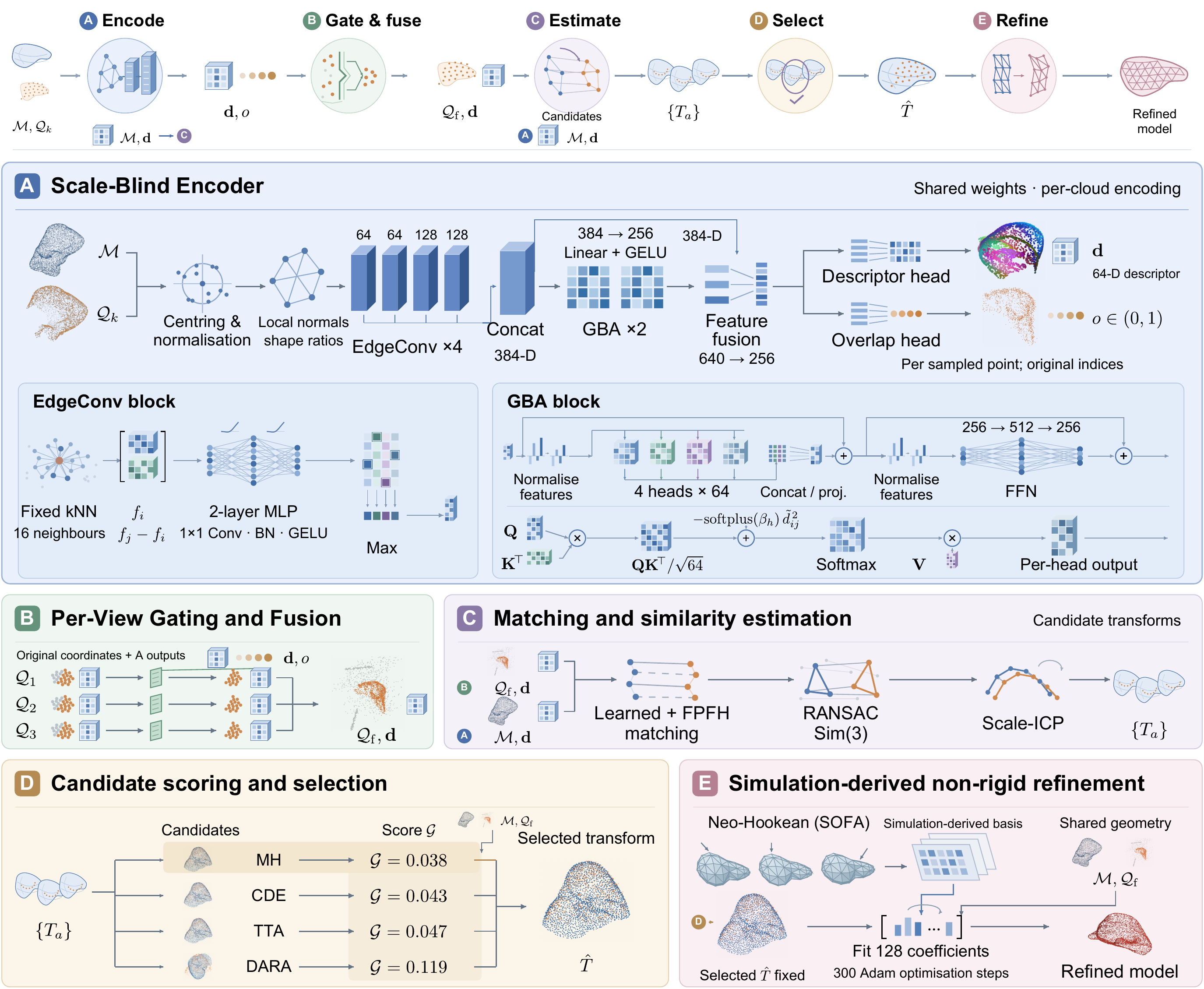}
\caption{SimCompose. The blue cloud $\Model$ samples the preoperative liver surface; the orange cloud $\Obs_k$ is the observation from view $k$. (A)~Each cloud is encoded independently to give descriptors $\desc$ and probabilities $o$ (model visibility or observation liver membership) for its sampled points. Descriptor colours show a principal-component projection; blue tiles and gold dots denote descriptors and probabilities. (B)~Point indices retained by per-view gating, a threshold on each view's probabilities $o$, select the original coordinates and descriptors for fusion into the gated observation $\Obsf$. (C)~Candidate similarity transformations $T_a$ are estimated, with $a$ indexing candidates. (D)~The geometric score $\gscore$, a trimmed observation-to-model surface distance, selects $\hat T$. (E)~With that transformation fixed, deformation is fitted in a simulation-derived subspace. Lettered ports connect reused outputs; repeated cloud symbols denote the same input coordinates.}
\label{fig:method}
\end{figure*}

SimCompose encodes surfaces independently and filters each view before fusion. It estimates candidate transformations from feature matches and selects one by surface agreement; disagreement between candidates supports failure screening. Optional non-rigid fitting then adjusts residual surface mismatch with the selected transformation fixed.

Let $\Model\subset\Real^3$ be the source cloud sampled from the preoperative liver surface and $\Obs_k$ the target cloud from the segmented intraoperative surface in view $k$. For $K$ views sharing coordinates and relative scale, we estimate the model-to-observation transformation
\begin{equation}\label{eq:prob}
  T(\p)=s\R\p+\tvec,\qquad \q\approx T(\p).
\end{equation}
Here $\p\in\Model$, $\q\in\Obs_k$, $s>0$ is uniform scale, $\R\in\mathrm{SO}(3)$ a $3\times3$ proper rotation, and $\tvec\in\Real^3$ translation. Together they define $T\in\Sim$, the similarity group. Surface reconstruction and segmentation precede registration. Only the encoder is trained offline, with synthetic supervision; inference uses neither anatomical landmarks nor true registration transforms.

\subsection{Scale-Blind Encoder}\label{sec:encoder}

Each cloud $X=\{\x_i\}_{i=1}^{N}$ is centred at $\bar{\x}$ and divided by its root-mean-square (RMS) distance from that centroid:
\begin{equation}\label{eq:norm}
  \tilde{\x}_i=(\x_i-\bar{\x})\Big/\sqrt{\tfrac{1}{N}\textstyle\sum_j\|\x_j-\bar{\x}\|^2},
\end{equation}
where the denominator defines $\rmsr(X)$. Independent normalisation removes absolute size from feature learning; the encoder is thus scale-blind, and deployment scale is not limited to the sampled training range. Original coordinates remain attached to point indices for later scale, rotation and translation estimation. Distance tolerances are scaled by $\rmsr(\Obsf)$, the RMS radius of the gated, fused observation (Section~\ref{sec:reg}).

A fixed $16$-nearest-neighbour (kNN) graph supplies local principal component analysis (PCA) normals and three normalised eigenvalue ratios, describing surface orientation and variation along the principal directions. These features join $\tilde{\x}_i$ before four edge-convolution (EdgeConv) blocks~\cite{wang2019dgcnn}, which update each point from features of its graph edges; all four reuse this fixed graph. Two geometry-biased attention (GBA) blocks then combine within-cloud feature similarity with a distance penalty:
\begin{equation}\label{eq:attn}
  A^{(h)}_{ij}=\frac{(W_Q^{(h)}\!f_i)^{\!\top}(W_K^{(h)}\!f_j)}{\sqrt{d_{\mathrm{h}}}}
  -\softplus(\beta_h)\,\tilde d_{ij}^{\,2},
\end{equation}
where $A^{(h)}_{ij}$ is the attention score of head $h$ between points $i$ and $j$, $f_i$ is the block input, $W_Q^{(h)},W_K^{(h)}$ are query/key projections (Fig.~\ref{fig:method} shows the resulting features $\mathbf Q,\mathbf K$ and values $\mathbf V$), $d_{\mathrm h}=64$ is the head dimension, and $\tilde d_{ij}=\|\tilde\x_i-\tilde\x_j\|$ is computed without gradients. The learned $\beta_h\in[-4,4]$ sets a nonnegative penalty through $\softplus(\beta_h)=\log(1+e^{\beta_h})$, the only positional encoding. Encoding is independent across clouds, with no cross-attention, so features can be cached before correspondence search.

A fused $256$-D feature yields a $64$-D, $L_2$-normalised descriptor $\desc_i\in\Sph^{63}$ and, from an overlap head, probability $o_i\in(0,1)$ at each input point index. Matching uses cosine similarity $S_{ij}=\desc_i^{\top}\desc_j$. Source probabilities describe visibility, since the complete model includes unobserved regions; target probabilities separate liver points from clutter and are the only probabilities used for gating, so the model remains ungated. The Complementary Descriptor Experts (CDE) encoder variant has local, neighbourhood and global descriptor heads.

\emph{Implementation.} We sample $2048$ model points and $1536$ points per view. EdgeConv widths are $64,64,128,128$ for each block using a multilayer perceptron (MLP), batch normalisation (BN) and Gaussian error linear units (GELU). Each GBA block uses four $64$-D heads and a feed-forward network (FFN). The base encoder has $1.52$\,M parameters.

\subsubsection{Synthetic Training Data}\label{sec:synthdata}

Each synthetic pair draws a mesh from the MedShapeNet collection of medical shapes~\cite{li2023medshapenet} or the P2I-LReg-derived training pool with probabilities $0.60$ and $0.40$. MedShapeNet supplies $1{,}107$ meshes: $850$ for encoder training and $257$ for evaluation, rescaled to a $0.30$\,m diagonal so $3$\,mm is organ-relative. The second pool derives from $21$ preoperative P2I-LReg meshes~\cite{zhou2025selfp2ir}, flattened along the PCA thickness direction; a point z-buffer extracts their visible surfaces by comparing point depths after projection. Training uses the preoperative anatomy of evaluation patients 01 and 02; neither training nor warm-start initialisation uses that of patients 13, 17 or 18. No evaluation patient's intraoperative cloud or real registration label is used.

Front-face cropping, which keeps only source points on the surface facing a virtual camera, targets a visible fraction of $0.40$--$1.0$ of source points. Displacement fields from SOFA simulations using a Neo-Hookean hyperelastic model~\cite{faure2012sofa,marchesseau2010fast} supply deformation augmentation and the offline refinement basis. Independent resampling reflects different mesh and reconstructed-surface sampling, so target and source indices are not correspondences. The resampled target points are placed in the observation frame by a known $\Sim$: log-uniform scale $[0.8,1.3]$, uniform rotation under the Haar measure on $\mathrm{SO}(3)$ and translation of $1.3$--$4$ organ diameters, where diameter denotes the $0.30$\,m model diagonal. Noise along viewing rays and clutter from background, abdominal wall, instruments and segmentation leakage are added. Multi-view pairs share one organ and transform across $K$ views.

As supervision, the known transform provides correspondence labels rather than a pose-regression target: source points, deformed if augmented, are mapped by it. With $\sigma$ the median nearest-neighbour distance from these points to the target, pairs below $1.5\sigma$ are positive and those above $4\sigma$ negative. The intervening band is ignored because independent resampling can place a true surface correspondence between sampled points. Source visibility labels use sampled geometry and distance thresholds; target labels distinguish liver-surface points from added clutter.

\subsubsection{Training Objective}\label{sec:training}

Circle loss~\cite{sun2020circle} encourages similar descriptors for positive pairs and separates negative pairs, using margin $0.1$ and scale $24$. Binary cross-entropy supervises source visibility and target liver-membership probabilities, so the objective sums descriptor loss and both unit-weight cross-entropy terms. During the first $3{,}000$ steps, occlusion, clutter and noise increase progressively. The AdamW optimiser runs for $20{,}000$ steps of $8$ pairs, using learning rate $10^{-3}$, weight decay $10^{-4}$, cosine decay and gradient clipping at $1.0$. CDE is then initialised from the base encoder and trained for a further $15{,}000$ steps.

\subsection{Per-View Gating and Fusion}\label{sec:reg}

Because clutter varies across views and the overlap head was trained on individual observations, we filter each view before fusion. Points are retained by a $64$-bin Otsu threshold~\cite{otsu1979threshold} (the value maximising between-class variance) on $o_i$, clipped to $[0.4,0.8]$, or by a threshold of $0.05$ when fewer than $8\%$ of the view's points have $o_i<0.25$. If thresholding retains fewer than $200$ points for a single pair, or fewer than $40$ per multi-view observation, the highest-scoring points are kept up to that minimum. Retained indices select both descriptors and original coordinates for fusion into $\Obsf$, with no further gating. Fusion relies on the reconstruction for consistent per-view depth scales and offsets.

\subsection{Matching and \texorpdfstring{$\Sim$}{Sim(3)} Estimation}\label{sec:corr}

Each retained target descriptor is matched to its nearest descriptor in the ungated model, provided the match is mutual and the nearest-to-second-nearest distance ratio does not exceed $0.97$ for learned descriptors or $0.95$ for fast point feature histograms (FPFH)~\cite{rusu2009fpfh}. FPFH uses neighbourhoods of $16$ and $40$ points. If fewer than six matches survive, both tests are dropped in favour of plain nearest-descriptor matching. Learned and FPFH matches are then pooled as putative correspondences, whose geometric consistency is tested next.

Random sample consensus (RANSAC)~\cite{fischler1981ransac} draws three-match samples, rejects samples whose largest target-to-source edge-length ratio exceeds the smallest by over $25\%$, and solves accepted samples in closed form~\cite{umeyama1991least}. After $6000$ iterations, or $4000$ per restart of multi-hypothesis RANSAC (Section~\ref{sec:select}), the estimate with the largest consensus at distance threshold $0.06\,\rmsr(\Obsf)$ is refitted on that consensus set. When fewer than four pairs or eight inliers remain, an ungated FPFH registration replaces this estimate only if it yields more inliers. Point-to-plane iterative closest point with scale estimation (scale-ICP)~\cite{chen1992object,besl1992icp} then uses $55$ iterations and the closest $92\%$ of observation-to-model pairs, excluding unobserved model regions.

\subsection{Candidate Scoring and Selection}\label{sec:select}

Smooth liver surfaces can support incorrect high-inlier alignments. We compare candidates $T_a=(s_a,\R_a,\tvec_a)\in\Sim$, indexed by $a$, obtained by complementary estimators on the same $\Model,\Obsf$.

Multi-hypothesis RANSAC (MH) uses three restarts. Each pose $(s,\R,\tvec)$ is expanded into four candidates by composing $\R$ with the identity or a $180^\circ$ rotation about a coordinate axis (large-rotation flips); the new rotation $\R'$ takes translation $\tvec'=\tvec+s(\R-\R')\bar\p$, which keeps the transformed model centroid $\bar\p$ in place, and the best three candidates receive iterative-closest-point refinement. The CDE variant supplies a separate match set from each descriptor head and returns the pose with the lowest geometric score $\gscore$ of~\eqref{eq:geom}.

Test-time adaptation (TTA) updates a small trainable adapter module of rank $16$ on frozen features for three rounds of five steps of the Adam optimiser. Each round pairs observed points with their nearest model points under the current transformation. The adapter brings their descriptors closer and penalises source-descriptor changes with weight $0.5$. Because TTA relies on correspondences from the current pose, a new transformation is accepted only if it improves the geometric score.

Deformation-aware rigid alignment (DARA) fits $18$ smooth deformation modes derived from the lowest-frequency eigenvectors of a $12$-neighbour graph Laplacian. The seven directions corresponding to local linear changes in rotation, translation and scale are removed first. DARA keeps the RANSAC scale and discards the fitted deformation before returning the transformation. Methods that search $\Sim$ by surface distance alone are excluded from the candidate sets, because the selection score could favour them.

For sorted distances $d_{(i)}$ from observed points to the transformed model, we select $\hat T=T_{\hat a}$ with $\hat a=\argmin_a\gscore(T_a)$:
\begin{equation}\label{eq:geom}
  \gscore(T)=\median\{\,d_{(1)},\dots,d_{(\lceil 0.8|\Obsf|\rceil)}\,\}\big/\rmsr(\Obsf),
\end{equation}
where the median uses the smallest $80\%$ of distances. The score evaluates surface agreement independently of descriptor inlier counts. Its observation-to-model direction accounts for partial visibility: reversing the direction would also penalise unobserved parts of the complete model. On Blender, candidates are instead selected by median silhouette Dice across the calibrated, segmented anchor views.

For screening, each candidate pair $(a,b)$ has disagreement
\begin{equation}\label{eq:dis}
  \delta_{ab}=\frac{\angle(\R_a^{\!\top}\R_b)}{0.05}
  +\frac{\|\tvec_a-\tvec_b\|}{0.10\,\bar s_{ab}}
  +\frac{|\log(s_a/s_b)|}{0.01},
\end{equation}
where $\angle(\cdot)$ is the rotation angle in radians and $\bar s_{ab}=(s_a+s_b)/2$. For the two-method set (MH and CDE), disagreement $\dis$ is the single $\delta_{ab}$; for the full candidate set (Section~\ref{sec:setup}), it is the median of $\delta_{ab}$ over pairs. Registrations with high $\dis$ can be withheld.

\subsection{Non-Rigid Refinement From Simulations}\label{sec:nonrigid}

In porcine studies, pneumoperitoneum, \ie gas insufflated into the peritoneal cavity for laparoscopy, displaces tissue by $7$--$19\%$ of organ extent~\cite{wise2024intraoperative,zijlmans2012navigated}. A similarity transformation cannot represent this tissue deformation. A partial, noisy surface, however, constrains an unrestricted warp only weakly. We therefore restrict refinement to a subspace derived from SOFA simulations~\cite{faure2012sofa,marchesseau2010fast} using a Neo-Hookean model with gravity, fixed ligament patches and surface loads. Young's modulus is $2.5$--$18$\,kPa and Poisson's ratio $0.45$--$0.48$; dynamic relaxation, \ie damped time integration until motion settles, replaces a static-equilibrium solve. Element-inversion and displacement checks retain $723$ fields from $372$ geometries with mean displacement $8$--$60$\,mm.

Each field is sampled on a $10^3$-node grid in its mesh's centred, PCA-aligned frame with RMS-radius normalisation. Geometry and displacement share the scaling to a $0.30$\,m reference diagonal. Singular value decomposition gives mean $\boldsymbol\mu$ and rank-$128$ basis $\mathbf B$ whose rows are displacement modes. The mean and modes are interpolated to each input model.

The online fit keeps $\hat T$ and the basis fixed to separate residual surface fitting from scale and pose search. We fit $128$ coefficients $\mathbf c\in\Real^{128}$ using $300$ Adam iterations at learning rate $0.05$, with no magnitude penalty or network training:
\begin{equation}\label{eq:nr}
  \mathbf{c}^{\star}=\argmin_{\mathbf{c}}\;\frac{1}{|\mathcal{I}|}
  \sum_{i\in\mathcal{I}}\;\min_{\mathbf{m}\in\widetilde{\Model}(\mathbf{c})}\lVert\q_i-\mathbf{m}\rVert,
\end{equation}
where $\widetilde\Model(\mathbf c)$ is deformed by $\boldsymbol\mu+\mathbf B^{\!\top}\mathbf c$ then posed by $\hat T$; $\q_i$ are fitting points and $\mathcal I$ retains their smallest $90\%$ of distances.

A fixed-seed $50/50$ split uses half the observed points for fitting and the other half for held-out evaluation; every initial transformation receives the same split and refinement procedure. Because smooth fields can also absorb pose error, we report the fitted displacement as a percentage of diameter alongside held-out surface distance.
\section{Experiments}\label{sec:experiments}

\subsection{Setup}\label{sec:setup}

\subsubsection{Domains, Metrics and Statistics}

We evaluate our method on held-out synthetic geometry, seven static Blender cases and five P2I-LReg patients. The in-vivo set contains $22$ keyframes from patient 01, $20$ from 02, $21$ from 13, $11$ from 17 and $18$ from 18. Point clouds are reconstructed using zero-shot Depth-Anything-3 depth estimates, case-specific camera intrinsics and liver masks.

For Blender, each of the $7$ cases is evaluated over $6$ folds, each with its own three anchor and eight held-out frames, and $3$ seed rotations of the initial pose, giving $126$ registrations per method. Evaluation frames may recur across folds, and views within a case share anatomy and the same reconstruction. We therefore use the case, rather than individual frames, as the statistical unit.

Each model is rescaled to a bounding-box diagonal of $300$\,mm, and all registration distances are reported at this reference scale. A synthetic registration is successful under our strict criteria only if it meets all three thresholds: i) relative translation error $\le3$\,mm; ii) relative rotation error $\le0.57^\circ$ between the estimated and true transformations; and iii) log-scale error $|\log(\hat s/s)|\le0.5\%$ for estimated and true scales $\hat s$ and $s$.

At a reference radius of $0.15$\,m, the rotation and scale limits correspond to displacements of about $1.5$ and $0.75$\,mm, respectively. Adding the translation limit gives a worst-case displacement budget of about $5.25$\,mm. These research criteria are motivated by published guidance targets~\cite{thompson2018invivo,cash2007concepts}.

Because ground-truth registration transforms are unavailable for the in-vivo and Blender data, we evaluate surface agreement using silhouette Dice and one-directional $90\%$-trimmed chamfer distance (CD). We compute CD as the mean observed-to-model distance after removing the largest $10\%$ of residuals.

For Table~\ref{tab:real}, CD is computed using $12{,}000$ model points and $2048$ observed points, and results are reported as the mean of per-case medians. For non-rigid results, CD uses $6000$ model points and independent runs of the rigid solver, with results reported as the median of per-patient medians. We compare case medians with exact two-sided Wilcoxon signed-rank tests (nonparametric, paired).

A \emph{collapse} is a large-error failure: CD $>20$\,mm or Dice $<50$ in vivo, and Dice $<40$ on Blender; missing in-vivo poses also count as collapses. In vivo, rigid-stage collapses are counted before refinement and held-out collapses after it, on points not used for fitting; Blender collapses are measured in held-out views.

\subsubsection{Baselines and Selection}

We compare our method with Self-P2IR's rigid core~\cite{zhou2025selfp2ir}, Lepard~\cite{li2022lepard} and GeoTransformer~\cite{qin2023geotransformer}. We also include two reference methods. Coherent point drift (CPD)~\cite{myronenko2010cpd}, a probabilistic point-set registration method, is run with comparable computational effort (matched-effort CPD). Chamfer-$\Sim$ directly optimises a chamfer distance over $\Sim$.

For the rigid $\mathrm{SE}(3)$ baselines, we supply scale using the ratio of the observed cloud's RMS radius to that of the model (extent-ratio scale). For Self-P2IR's rigid core, this replaces its native principal-axis ratio.

In addition to Self-P2IR's rigid core, we evaluate its complete released system, including its own preprocessing, gating and neural-deformation-pyramid (NDP) fit. Self-P2IR uses released weights trained on the cohort's training split, whereas GeoTransformer and Lepard use public weights trained on indoor scans. We additionally retrain GeoTransformer on our synthetic data using two seeds (domain-retrained GeoTransformer).

On synthetic data, SimCompose selects from a full candidate set of five methods: MH, CDE, TTA, and two additional MH variants using encoders trained with two other seeds. In vivo and on Blender, the full candidate set has four methods: MH, CDE, TTA, and DARA.

Because CD and $\gscore$ both measure surface distance, we also select candidates on half the points and measure CD on the other half; this raises SimCompose's median of per-patient median CD by $0.48$\,mm.

\subsection{Comparative Evaluation}\label{sec:quant}

\subsubsection{In-Vivo Comparison}
SimCompose gives lower CD than the released learned methods in Table~\ref{tab:real}: $12.1$ versus $44.5$--$74.4$\,mm. Its Dice exceeds Self-P2IR's rigid core on every patient, with mean gains of $10.3$ points overall and $8.9$ on the three patients whose preoperative anatomy never enters encoder training. The gains are $18.6$, $1.7$ and $6.4$ points for patients 13, 17 and 18, respectively. Chamfer-$\Sim$ achieves higher Dice but higher CD. With five patients, the smallest attainable exact two-sided $p$ is $0.0625$, so no in-vivo comparison can reach $p<0.05$ and we report effect sizes.

For Self-P2IR's rigid core, the extent-ratio rule gives Dice $69.3$ and CD $45.9$\,mm, compared with Dice $70.1$ and CD $47.7$\,mm under its native principal-axis rule.

Applied to released GeoTransformer and Lepard correspondences, our solver still produces rigid-stage collapses on $73.9\%$ and $89.1\%$ of keyframes, respectively; with Lepard correspondences, it returns no pose on $71/92$ keyframes. Domain-retrained GeoTransformer has $91$--$96\%$ collapses with its native solver and $62$--$64\%$ with ours.

Even choosing each patient's best external scale multiplier from $0.7$--$2.0$ leaves GeoTransformer and Self-P2IR at median CD $33.7$ and $43.2$\,mm, versus $11.2$\,mm for SimCompose (Fig.~\ref{fig:quant}a). This multiplier is an evaluation-only oracle that minimises CD rather than measuring physical scale.

\subsubsection{Blender Comparison}
SimCompose achieves Dice $73.6$, compared with $66.1$ for Self-P2IR's rigid core, a gain of $7.5$ points. Its CD is $7.8\%$ versus $5.1\%$ of diameter. Chamfer-$\Sim$ has no held-out-view collapses in $126$ fits, yet passes only $2$--$2.7\%$ of synthetic strict pose tests.

\subsubsection{Synthetic Pose Accuracy}
Self-P2IR, Lepard, GeoTransformer and PREDATOR~\cite{huang2021predator} each pass $0/300$ strict tests with either extent-ratio or exact scale (Table~\ref{tab:synth}). SimCompose reaches $77\%$ with three views and candidate selection. Applying our $\Sim$ solver to GeoTransformer or Lepard correspondences raises each to $15.3\%$, but neither passes the clutter subset, the $150$ of these tests with added clutter ($0/150$).

GeoTransformer retraining follows its released training recipe for 3DMatch~\cite{zeng20173dmatch}, an indoor-scan registration benchmark, and runs for $40$ epochs of $12{,}000$ pairs, with rigid-only pairs and true-scale alignment during training. Evaluation still uses external extent-ratio scale. Although both retraining runs converge, the native solver passes $0/300$ strict tests: median rotation error is $3.4$--$4.0^\circ$ and scale error remains $13\%$. Our solver raises retrained success to $19.7$--$20.7\%$, versus $15.3\%$ before retraining.

For comparison with registration benchmarks, we re-evaluate the same test pairs under coarser rotation and translation tolerances, first without and then with relaxed scale limits. At organ-rescaled 3DMatch criteria ($15^\circ$, $10\%$ of diameter)~\cite{choy2020deep,zeng20173dmatch} and criteria of the outdoor laser-scan benchmark KITTI ($5^\circ$, $2\%$)~\cite{yew20183dfeat,qin2023geotransformer}, without a scale constraint, SimCompose passes $98\%$ at both tiers. Retrained GeoTransformer's native solver reaches $90$--$92\%$ and $35\%$. Adding scale limits of $5\%$ and $1\%$ reduces it to $28$--$29\%$ and $7\%$, while SimCompose reaches $98\%$ and $97\%$.

\subsubsection{Non-Rigid Refinement and Complete-System Control}
The simulation-derived fit reduces held-out CD to $9.9$\,mm for SimCompose, with displacement $5.0\%$ of diameter (Table~\ref{tab:nonrigid}). Patient-median CD is $9.9$\,mm for patient 01, $8.6$\,mm for 02, $10.2$\,mm for 13, $10.4$\,mm for 17 and $8.1$\,mm for 18. Rigid baselines also reach $10.7$--$12.9$\,mm after fitting, but require $12$--$13\%$ displacement. NDP~\cite{li2022ndp} obtains lower error with $34.3\%$ displacement from an $88.3$\,mm initial error. The static Blender organ, with zero true deformation, yields $6.1\%$ fitted displacement. Equal-rank generic bases (random smooth fields and graph-Laplacian modes) give $10.2$ and $10.1$\,mm, respectively, versus $9.9$\,mm.

Self-P2IR's complete released system is tested on its own five-patient test split, whose patients were not used in its real-data tuning. Median CD improves from $71.7$ to $67.0$\,mm after its NDP fit, with Dice $29.0$ and displacement $6.6\%$. Its rendered-silhouette protocol gives Dice $35.6\pm15.9$. Its rigid core under our protocol gives $48.8$\,mm by the same cross-patient statistic. This control points to rigid initialisation as a limitation of Self-P2IR's complete system.

\begin{table}[t]
\caption{Rigid registration on in-vivo and Blender data.}
\label{tab:real}
\TableSetup
\begin{tabularx}{\linewidth}{@{}Llrrrr@{}}
\toprule
 & & \multicolumn{2}{c}{\textbf{In vivo}} & \multicolumn{2}{c}{\textbf{Blender}}\\
\cmidrule(lr){3-4}\cmidrule(lr){5-6}
\TableHead{Method} & \TableHead{Scale} & \TableHead{Dice~$\uparrow$} & \TableHead{CD~$\downarrow$\\(mm)} & \TableHead{Dice~$\uparrow$} & \TableHead{CD~$\downarrow$\\(\%)}\\
\midrule
SimCompose (ours) & blind & \textbf{79.6$\pm$6.0} & \textbf{12.1$\pm$1.8} & \textbf{73.6} & 7.8\\
Self-P2IR~\cite{zhou2025selfp2ir}, rigid core$^\dagger$ & ext.-ratio & 69.3$\pm$6.2 & 45.9$\pm$11.6 & 66.1 & \textbf{5.1}\\
Lepard~\cite{li2022lepard} & ext.-ratio & 66.0$\pm$5.2 & 44.5$\pm$5.3 & 62.4 & 5.5\\
GeoTransformer~\cite{qin2023geotransformer} & ext.-ratio & 56.3$\pm$4.4 & 74.4$\pm$18.8 & 48.3 & 9.5\\
GeoTr., retrained$^{\mathrm{r}}$ & ext.-ratio & 63.8 & 63.4 & --- & ---\\
GeoTr., retr.\ $+$ our solver$^{\mathrm{r,s}}$ & pre-scaled & 75.6 & 73.7 & --- & ---\\
\textit{Chamfer-}$\Sim$ & blind & \textit{82.9$\pm$3.7} & \textit{17.7$\pm$3.9} & \textit{80.4} & \textit{3.8}\\
\textit{CPD, matched-effort}~\cite{myronenko2010cpd} & clamped & \textit{40.7$\pm$4.1} & \textit{64.9$\pm$17.4} & \textit{72.3} & \textit{3.9}\\
\bottomrule
\end{tabularx}
\TableNotes{Means over $5$ in-vivo patients and $7$ Blender cases; $\pm$ denotes cross-patient standard deviation where available. Blender CD is \% of diameter. Bold: best non-reference entry; italics: reference methods and oracle bounds; ---: not run or not applicable. GeoTr.: GeoTransformer; retr.: retrained. Scale is estimated internally (blind), supplied by the RMS-extent ratio (ext.-ratio), supplied before a new $\Sim$ solve (pre-scaled), limited to a window (clamped), or set by the original method (native). Scale labels, italics, dashes and, where used, bold follow this convention in all tables.
$^\dagger$Weights trained on this cohort's training split.
$^{\mathrm{r}}$Domain-retrained, two-seed mean.
$^{\mathrm{s}}$Mean strongly influenced by patient~17; cross-patient median $26.9$--$30.6$\,mm.}
\end{table}

\begin{table}[t]
\caption{Registration accuracy on held-out synthetic geometry.}
\label{tab:synth}
\TableSetup
\begin{tabularx}{\linewidth}{@{}Llrr@{}}
\toprule
\TableHead{Configuration} & \TableHead{Scale} & \TableHead{Views} & \TableHead{Pass rate\\(\%)~$\uparrow$}\\
\midrule
Self-P2IR / Lepard / GeoTr. / PREDATOR$^{\mathrm{a}}$ & ext.-ratio & 1 & $0$\\
GeoTr., retrained$^{\mathrm{r}}$ & ext.-ratio & 1 & $0$\\
\textit{Chamfer-}$\Sim$\textit{ (ref.)} & blind & 1 & $\mathit{2.7}$\\
FPFH/TEASER++/SC$^2$-PCR~\cite{rusu2009fpfh,yang2021teaser,chen2022sc2pcr} & blind & 1 & $\le 8$\\
GeoTr.\ $+$ our solver$^{\mathrm{b}}$ & pre-scaled & 1 & $15.3$\\
Lepard $+$ our solver$^{\mathrm{b}}$ & pre-scaled & 1 & $15.3$\\
GeoTr., retr.\ $+$ our solver$^{\mathrm{b,r}}$ & pre-scaled & 1 & $20.2$\\
\midrule
\textit{Chamfer-}$\Sim$\textit{ (ref.)} & blind & 3 & $\mathit{2}$\\
SimCompose (ours) & blind & 3 & $\mathbf{77}$\\
\textit{Oracle (upper bound)} & blind & 3 & $\mathit{89}$\\
\bottomrule
\end{tabularx}
\TableNotes{Pass rates use the strict accuracy thresholds in Section~\ref{sec:setup}. One view: $n{=}300$; three views: $n{=}100$. Oracle selects the ground-truth best candidate per pair for analysis only. TEASER++ and SC$^2$-PCR are classical outlier-robust registration solvers.
$^{\mathrm{a}}$Methods~\cite{zhou2025selfp2ir,li2022lepard,qin2023geotransformer,huang2021predator}: each run in its own software environment; all identical under exact scale.
$^{\mathrm{b}}$Correspondences re-solved by our $\Sim$ solver.
$^{\mathrm{r}}$Domain-retrained, two-seed mean; runs with our solver: $59/300$, $62/300$.}
\end{table}

\begin{table}[t]
\caption{Non-rigid refinement on in-vivo data.}
\label{tab:nonrigid}
\TableSetup
\begin{tabularx}{\linewidth}{@{}Llrrrr@{}}
\toprule
\TableHead{Method} & \TableHead{Scale} & \TableHead{CD$_\mathrm{r}$~$\downarrow$\\(mm)} & \TableHead{CD$_\mathrm{nr}$~$\downarrow$\\(mm)} & \TableHead{Dice~$\uparrow$} & \TableHead{Def.\\(\%)}\\
\midrule
SimCompose (ours) & blind & 14.2 & 9.9 & 83.4 & 5.0\\
NDP~\cite{li2022ndp}$^\ddagger$ & blind & 88.3 & 6.0 & 89.0 & 34.3\\
Lepard~\cite{li2022lepard} & ext.-ratio & 45.4 & 10.7 & 83.3 & 12.1\\
Self-P2IR~\cite{zhou2025selfp2ir}, rigid core$^\dagger$ & ext.-ratio & 48.8 & 11.2 & 82.9 & 12.8\\
Self-P2IR, complete system$^{\mathrm{e}}$ & native & 71.7 & 67.0 & 29.0 & 6.6\\
GeoTransformer~\cite{qin2023geotransformer} & ext.-ratio & 82.5 & 12.9 & 73.0 & 12.5\\
\textit{Chamfer-}$\Sim$ & blind & \textit{17.1} & \textit{8.8} & \textit{89.3} & \textit{6.6}\\
\textit{CPD, matched-effort}~\cite{myronenko2010cpd} & clamped & \textit{63.7} & \textit{11.3} & \textit{67.6} & \textit{24.0}\\
\bottomrule
\end{tabularx}
\TableNotes{Medians over $5$ patients. CD$_\mathrm{r}$/CD$_\mathrm{nr}$ are held-out CD before/after fitting; Dice is after fitting; Def.\ is fitted displacement as \% of diameter. Unmarked rows use the identical simulation-derived SOFA subspace.
$^\ddagger$NDP uses its own neural pyramid.
$^{\mathrm{e}}$Complete released Self-P2IR system, end-to-end on its own test split with our metrics.
CD$_\mathrm{r}$ uses sparser sampling than Table~\ref{tab:real}; $^\dagger$ as in Table~\ref{tab:real}.}
\end{table}

\subsubsection{Qualitative Results}
\begin{figure*}[t]
\centering
\includegraphics[width=\textwidth]{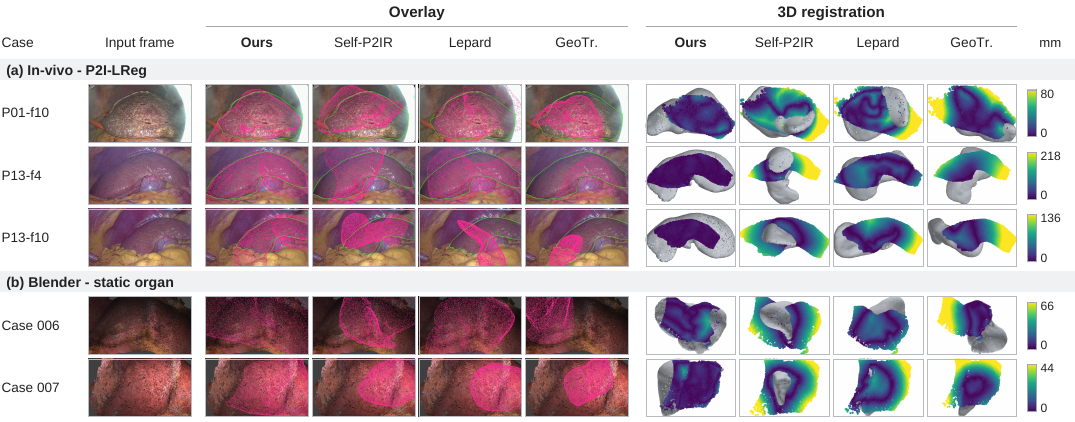}
\caption{Rigid-stage comparisons. \textbf{Ours}: SimCompose; GeoTr.: GeoTransformer. (a)~P2I-LReg keyframes; Pnn-fm denotes patient nn, frame m. (b)~Blender cases: static organ, zero true deformation. Columns: input, overlays, and 3D registration. Overlay models are magenta; boundaries are green in (a). Registration models are grey; observations use viridis residual colours, dark $=$ low. Each row shares a camera and colour scale, in mm at right; maxima are the four-method 95th percentiles, with larger residuals saturated. Non-rigid stage: Fig.~\ref{fig:elastic}.}
\label{fig:qual-invivo}
\end{figure*}

\begin{figure}[t]
\centering
\includegraphics[width=\columnwidth]{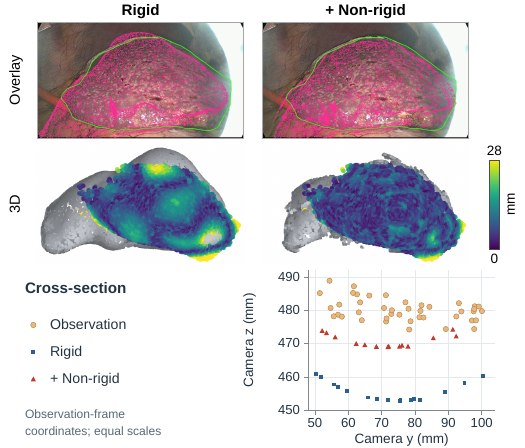}
\caption{Non-rigid refinement of P01-f10. Overlays and 3D registrations share a camera and colour scale in mm; colours follow Fig.~\ref{fig:qual-invivo}. Mean residual decreases from $10.3$ to $6.7$\,mm. Bottom: an $18$\,mm-thick slab parallel to the camera $y$--$z$ plane, centred on the observed point whose rigid residual is closest to the $97$th percentile. Observations are light discs, rigid points blue squares, and non-rigid points red triangles. Fitting narrows the gap between observations and model points.}
\label{fig:elastic}
\end{figure}

Fig.~\ref{fig:qual-invivo} compares rigid-stage size and position errors across methods, and Fig.~\ref{fig:elastic} shows refinement reducing residuals on one keyframe.

\begin{figure*}[t]
\centering
\includegraphics[width=\textwidth]{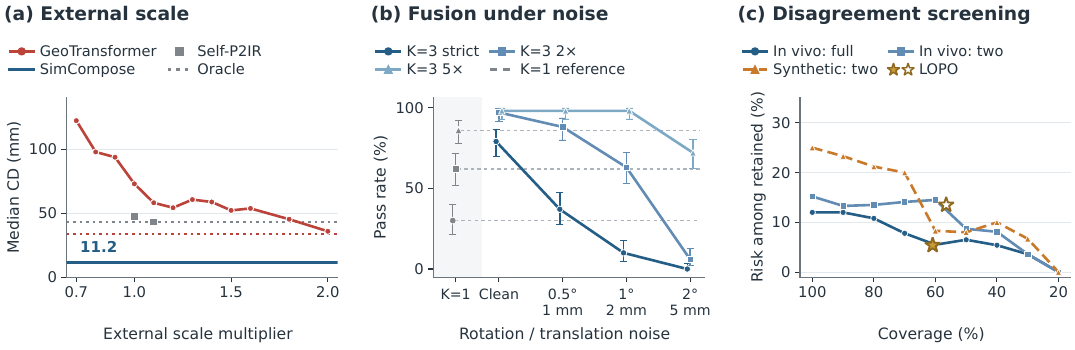}
\caption{Quantitative analyses. (a)~CD is the median of five patient medians, in millimetres at the reference scale. Dotted lines show CD at per-patient oracle multipliers: $33.7/43.2$\,mm for GeoTransformer/Self-P2IR; SimCompose estimates scale internally. Squares use shared multipliers $1.0/1.1$. (b)~Three-view fusion uses the full synthetic candidate set and $100$ pairs per point; clean strict success is $79\%$, as in Table~\ref{tab:ablation}. Error bars show two-sided $95\%$ Clopper--Pearson (exact binomial) intervals; grey points and lines are single-view references on the same pairs. The $2\times/5\times$ criteria multiply all strict thresholds in Section~\ref{sec:setup}. (c)~Curves show within-sample ranking at nominal coverage, the fraction of registrations retained after screening; stars show achieved coverage under leave-one-patient-out (LOPO) calibration. Synthetic events are strict-criterion failures; in-vivo events are rigid-stage collapses. Filled/open stars denote full/two-method sets.}
\label{fig:quant}
\end{figure*}

\subsection{Robustness and Selective Prediction}\label{sec:robust}

\subsubsection{Collapse Rates}
The single-method controls use one encoder and one registration pipeline without selection, on one cloud per in-vivo keyframe or three fused anchor views on Blender.

In-vivo rigid-stage collapses fall from $48/92$ for one method to $11/92$ for the full candidate set. After identical refinement, held-out collapses fall from $14/92$ ($15.2\%$) for one method to $1/92$ ($1.1\%$) for the full set. Blender held-out collapses fall from $23.0\%$ to $8.7\%$. Self-P2IR's rigid core ($0.0/6.3\%$ in vivo/Blender) and Lepard ($1.1/4.8\%$) collapse no more often under these held-out criteria, despite poorer in-vivo rigid CD, so held-out collapse rates alone do not identify a good rigid pose.

\subsubsection{Latency}
Single-GPU latency on NVIDIA RTX 5090 is $2.17$\,s per registration for the full candidate set, $1.73$\,s for the two-method set, $0.64$\,s for one method and $0.28$\,s for non-rigid fitting.

\newcommand{\rlab}[1]{\rotatebox[origin=c]{90}{\tiny #1}}
\newcommand{\pnl}[3]{\parbox[c][#2][c]{#1}{\centering%
  \includegraphics[width=#1,height=#2,keepaspectratio]{results/#3}}}
\newcommand{\gcell}[3]{{\setlength{\fboxsep}{0pt}\setlength{\fboxrule}{0.2pt}%
  \fcolorbox{black!30}{white}{\includegraphics[width=#1,height=#2]{grid4/#3}}}}
\newcommand{\gP}[1]{\gcell{1.656cm}{0.93cm}{#1}}
\newcommand{\gR}[1]{\gcell{1.442cm}{0.93cm}{#1}}
\newcommand{\gC}[1]{\begin{tabular}[c]{@{}c@{}}%
  {\tiny #1}\\[-0.5pt]
  {\setlength{\fboxsep}{0pt}\setlength{\fboxrule}{0.2pt}%
   \fcolorbox{black!30}{white}{\includegraphics[width=0.13cm,height=0.50cm]{grid4/cbar_strip.png}}}\\[-0.5pt]
  {\tiny 0}\end{tabular}}
\newcommand{\gCe}[1]{\begin{tabular}[c]{@{}c@{}}%
  {\tiny #1}\\[-0.5pt]
  {\setlength{\fboxsep}{0pt}\setlength{\fboxrule}{0.2pt}%
   \fcolorbox{black!30}{white}{\includegraphics[width=0.15cm,height=1.30cm]{grid4/cbar_strip.png}}}\\[-0.5pt]
  {\tiny 0}\end{tabular}}
\newcommand{\pE}[1]{\pnl{4.05cm}{1.64cm}{#1}}
\newcommand{\pEc}[1]{\pnl{4.05cm}{2.76cm}{#1}}
\newcommand{\pnlN}[3]{\parbox[c][#2][c]{#1}{\centering%
  \includegraphics[width=#1,height=#2,keepaspectratio]{results_new3d/#3}}}
\newcommand{\pQn}[1]{\pnlN{1.70cm}{0.94cm}{#1}}
\newcommand{\pBn}[1]{\pnlN{1.70cm}{1.34cm}{#1}}
\newcommand{\pEn}[1]{\pnlN{4.05cm}{1.70cm}{#1}}

\subsubsection{Selective Prediction}
Selective prediction withholds registrations ranked as likely failures; screening ranks them by candidate disagreement, without true registration transforms. In $300$ single-view synthetic tests, retaining $20\%$ raises strict success from $41.3\%$ to $61.7\%$. In the two-method multi-view test, disagreement detects strict failures with area under the receiver-operating-characteristic curve (AUROC) $0.742$; $25/100$ poses fail. The prespecified synthetic threshold $\dis^*=0.297$ retains $5.4\%$ of in-vivo frames, where disagreement is two orders of magnitude higher than on synthetic data.

For in-vivo leave-one-patient-out calibration, the threshold is the prespecified $60\%$ quantile of scores from the other four patients (Fig.~\ref{fig:quant}c). The full set retains $56/92$ frames with rigid-stage collapse risk $3/56=5.4\%$, versus $12.0\%$ unscreened. The one-sided $95\%$ binomial upper confidence bound on this risk is $13.3\%$ (frames treated as independent); the three retained collapses come from one patient, and per-patient coverage is $27$--$81\%$. The two-method set retains $57\%$ at $13.5\%$ rigid-stage risk versus $15.2\%$ unscreened.

\subsubsection{Inter-View Errors and Internal Targets}
To test sensitivity to inconsistent views, we perturb camera extrinsics and inter-frame geometry. At $K=3$ with $100$ pairs per noise level, $0.5^\circ/1$\,mm extrinsic noise gives $37\%$ strict success versus $30\%$ for one view of the same pairs. At $1^\circ/2$\,mm, the $2\times$-threshold rates are $63\%$ versus $62\%$. At $2^\circ/5$\,mm, all fused rates fall below single-view references (Fig.~\ref{fig:quant}b). A warp of $2\%$ of view extent reduces strict success to $1\%$.

Synthetic target-registration error (TRE) measures the distance from an estimated subsurface target to its known position. Across $100$ multi-view pairs, the distribution of per-pair median TRE has a median of $0.47$\,mm and a $90$th percentile of $2.0$\,mm. The correlation between trimmed surface CD and TRE is weak (Pearson $r=0.18$). Two pairs have TRE values of $27.5$ and $58.3$\,mm despite trimmed CD values of only $9$--$10$\,mm.

\subsection{Ablation Study}\label{sec:ablation}
The synthetic single-method control uses the base encoder with one correspondence pass and one RANSAC run on a single view. It gives $38.4\%$ success, averaged over three training seeds.

The multi-view ablations share one set of runs, in which the full model reaches $79\%$ ($77\%$ in Table~\ref{tab:synth}; TTA adds run-to-run variation). Table~\ref{tab:ablation} attributes $9$ percentage points to selection over the best fixed method and $37.7$ to fusion over single-view candidates. Oracle selection of the ground-truth best candidate reaches $89\%$.

\emph{Retraining.} Retrained encoders (two scratch runs and one warm-started run) give three-seed means of $33.2$--$36.3\%$ single-view, single-method success, versus $38.4\%$ for the original encoders; their multi-view success is $79\%$, matching the full model in Table~\ref{tab:ablation}.

\subsubsection{Scale Estimation}
Fixing extent-ratio scale and solving only $\mathrm{SE}(3)$ lowers all methods and oracle selection to $2\%$. Median rotation, translation and scale errors are $24.1^\circ$, $36$\,mm and $14.4\%$. Even exact scale gives only $53\%$, because fixing scale changes pose search and hypothesis rejection.

\subsubsection{Other Components}
Adding Chamfer-$\Sim$ to the five-method synthetic candidate set, with selection by the lowest $\gscore$ as before, gives $75\%$ success, compared with $79\%$ for the full model.

Removing TTA or large-rotation flips changes full-model success by at most $2$ points, although the affected individual method loses up to $11$. Gating the fused cloud once rather than each view gives $77\%$ versus $79\%$ for the full model and lowers oracle success from $89\%$ to $83\%$.

The two-method set (MH and CDE) gives $75\%$ strict multi-view success versus $79\%$ for the full set. In vivo, it gives mean patient-median Dice $78.9$ and CD $13.2$\,mm, versus Dice $79.6$ and CD $12.1$\,mm for the full set. Rigid-stage collapses rise from $11/92$ to $14/92$, while held-out collapses remain $1/92$ on a different frame.

\begin{table}[t]
\caption{Ablation of registration components and supplied scale.}
\label{tab:ablation}
\TableSetup
\begin{tabularx}{\linewidth}{@{}Lrr@{}}
\toprule
\TableHead{Configuration} & \TableHead{Pass rate\\(\%)~$\uparrow$} & \TableHead{$\Delta$\\(pp)}\\
\midrule
Full model (fusion $+$ methods $+$ selection) & $\mathbf{79}$ & ---\\
\textit{Oracle (upper bound)} & $\mathit{89}$ & $\mathit{+10}$\\
\midrule
w/o candidate selection (best fixed method) & $70$ & $-9$\\
w/o multiple methods (MH only) & $59$ & $-20$\\
w/o multi-view fusion (combined candidates, \emph{sv}) & $41.3$ & $-37.7$\\
w/o fusion and multiple methods (\emph{sv}) & $38.4$ & $-40.6$\\
\midrule
supplied ext.-ratio scale, $\mathrm{SE}(3)$ solve & $2$ & $-77$\\
supplied \emph{exact} ground-truth scale, $\mathrm{SE}(3)$ solve & $53$ & $-26$\\
\bottomrule
\end{tabularx}
\TableNotes{Pass rates use the strict accuracy thresholds of Table~\ref{tab:synth}, with $K{=}3$ views and $n{=}100$ unless marked \emph{sv}: single-view, $n{=}300$. $\Delta$ is the difference from the full model in percentage points (pp); w/o means without. Multi-view rows come from the same ablation runs, in which the full model reaches $79\%$ ($77\%$ in Table~\ref{tab:synth}). Oracle selects the ground-truth best candidate for analysis only.}
\end{table}

\section{Discussion and Conclusion}\label{sec:discussion}

The results separate the contributions of three components. Joint scale estimation is the foundation: supplying extent-ratio scale to an $\mathrm{SE}(3)$ solve lowers synthetic success to $2\%$, and even exact scale reaches only $53\%$ (Table~\ref{tab:ablation}). Because the encoder never sees metric size, the same weights register non-metric monocular reconstructions, such as the zero-shot Depth-Anything-3 clouds used in vivo, without supplied scale. Multi-view fusion then adds $37.7$ points over single-view candidates, and candidate selection adds $9$ points over the best fixed method.

Candidate selection also applies to single views: on single in-vivo frames, it reduces rigid-stage collapse from $48/92$ for one method to $11/92$ for the full set. Self-P2IR provides separate evidence that scale matters: under its own five-fold cohort cross-validation and metric, removing scale-consistent recovery raises the reported chamfer error of its deformation network from $2.97$ to $10.66$\,mm~\cite{zhou2025selfp2ir}.

Candidate disagreement is a ground-truth-free reliability signal. It detects strict two-method synthetic failures with AUROC $0.742$, and leave-one-patient-out screening of the full set lowers in-vivo rigid-stage collapse risk from $12.0\%$ to $5.4\%$ while retaining $56$ of $92$ frames.

Training-free refinement in the shared simulation-derived subspace lowers the held-out CD of SimCompose from $14.2$ to $9.9$\,mm with $5.0\%$ displacement, whereas rigid baselines need $12$--$13\%$ displacement to reach $10.7$--$12.9$\,mm (Table~\ref{tab:nonrigid}).

In a surgical workflow, SimCompose needs only a segmented laparoscopic reconstruction and the preoperative liver surface for each frame. Registration with the full candidate set takes $2.17$\,s on a single GPU, optional refinement adds $0.28$\,s, and disagreement can flag unreliable overlays.

The rendered multi-view dataset complements both evaluations: its known camera geometry tests whether one registration stays consistent across held-out views, where SimCompose gains $7.5$ Dice points over Self-P2IR's rigid core.

Because the in-vivo and Blender data lack true registration transforms, guidance accuracy near a resection margin remains to be measured against an in-vivo target reference. Cohorts larger than the five patients evaluated here can quantify variation with anatomy and acquisition conditions.

Multi-view use requires reliable camera geometry, consistent relative scale and consistent anatomy; Section~\ref{sec:robust} quantifies sensitivity to extrinsic and inter-frame errors. A tracked phantom or porcine study could test reconstruction, tracking and registration together against independently measured targets.

\label{sec:conclusion}%
SimCompose achieves $77\%$ strict synthetic success with three views and candidate selection, versus $38.4\%$ for one view and one method. In vivo, its rigid CD is $12.1$ versus $44.5$--$74.4$\,mm for released learned baselines, and optional refinement reduces median held-out CD from $14.2$ to $9.9$\,mm. Pose accuracy, surface agreement and failure screening remain distinct outcomes that should be assessed together.

\bibliographystyle{IEEEtran}
\bibliography{refs}

\end{document}